\documentclass[letterpaper, 10 pt, journal]{IEEEtran}
\usepackage{cite}
\ifCLASSINFOpdf
  \usepackage[pdftex]{graphicx}
  \usepackage{graphicx}
  \usepackage{array}
  \usepackage{tabularx}
  \usepackage{booktabs}
  \usepackage{caption}
  \usepackage{url}
  \DeclareGraphicsExtensions{.pdf,.jpeg,.png}
\else
\fi
\usepackage{amsmath}
\usepackage{dsfont}
\usepackage{amsfonts} 
\usepackage{tablefootnote}
\usepackage{multirow}

\ifCLASSOPTIONcompsoc
 \usepackage[caption=false,font=normalsize,labelfont=sf,textfont=sf]{subfig}
\else
 \usepackage[caption=false,font=footnotesize]{subfig}
\fi
\begin{document}
% \title{Navigation in narrow environment using imitation learning: Simulation and experiment}
\title{LiCS: Navigation using Learned-imitation on Cluttered Space}

\author{Joshua~Julian~Damanik,
        Jae-Won~Jung,
        Chala~Adane~Deresa,
        and~Han-Lim~Choi% <-this % stops a space
\thanks{J. J. Damanik, J.-W. Jung, C. A. Deresa, and H.-L. Choi were with the Department
of Aerospace Engineering, Korea Advanced Institute of Science and Technology, Daejeon,
Republic of Korea. E-mail: (joshuad, jjwon13, czchal99, hanlimc)@kaist.ac.kr.}% <-this % stops a space
% \thanks{J. Doe and J. Doe are with Anonymous University.}% <-this % stops a space
% \thanks{Manuscript received April 19, 2005; revised August 26, 2015.}}
}


% \markboth{IEEE Robotics and Automation Letters, ~Vol.~14, No.~8, August~2024}%
% {Shell \MakeLowercase{\textit{et al.}}: Bare Demo of IEEEtran.cls for IEEE Journals}
\markboth{\textit{T\MakeLowercase{his work has been submitted to the} IEEE \MakeLowercase{for possible publication.} C\MakeLowercase{opyright may be transferred without notice, after which this version may no longer be accessible.}}%
}{}

% \IEEEpubid{This work has been submitted to the IEEE for possible publication. Copyright may be transferred without notice, after which this version may no longer be accessible.}

% \IEEEspecialpapernotice{(Invited Paper)}

\maketitle

\begin{abstract}
In this letter, we propose a robust and fast navigation system in a narrow indoor environment for UGV (Unmanned Ground Vehicle) using 2D LiDAR and odometry. We used behavior cloning with Transformer neural network to learn the optimization-based baseline algorithm. We inject Gaussian noise during expert demonstration to increase the robustness of learned policy. We evaluate the performance of LiCS using both simulation and hardware experiments. It outperforms all other baselines in terms of navigation performance and can maintain its robust performance even on highly cluttered environments. During the hardware experiments, LiCS can maintain safe navigation at maximum speed of $1.5\ m/s$.
\end{abstract}


% \fi
\IEEEpeerreviewmaketitle

\section{Introduction}


% for IEEE journal papers produced under \LaTeX\ using
% IEEEtran.cls version 1.8b and later.
% % You must have at least 2 lines in the paragraph with the drop letter
% % (should never be an issue)
% I wish you the best of success.



\IEEEPARstart{N}{avigation} within cluttered indoor environments poses a substantial challenge for Unmanned Ground Vehicles (UGVs). Ensuring robust and rapid navigation in such cluttered spaces is vital for applications that range from warehouse automation to search and rescue missions. Traditional navigation systems often encounter difficulties in these environments due to tight spaces and numerous obstacles \cite{fox1997dynamic, quinlan1993elastic}.

The recent availability of benchmarking datasets \cite{perille2020benchmarking, xia2020interactive} for navigation in cluttered environments has facilitated significant advancements in learning-based navigation systems, particularly through the use of reinforcement learning (RL) \cite{xu2023benchmarking} and imitation learning (IL) \cite{zhang2016query}. RL shows promise but can result in unexpected behaviors and requires extensive reward function engineering for effective training \cite{li2018deep}.

Imitation learning (IL), in contrast, aims to replicate the behavior of an expert, be it a human or optimal control algorithm. This complex task can be simplified into a supervised learning model known as Behavior Cloning (BC). However, BC faces limitations due to its assumption that data in the training dataset are sampled independently of the environment \cite{beygelzimer2005error}. In practice, actions taken during training influence future states, leading to compounded errors in the learned policy \cite{ross2010efcient}.

To mitigate these challenges, techniques like SMILe \cite{ross2010efcient} and DAgger \cite{ross2011reduction} combine BC with sequential online learning to maintain policy performance. Although effective, these methods require ongoing expert interaction during training, which can be resource-intensive. Offline BC, alternatively, emphasizes careful planning of the demonstrations to ensure comprehensive exploration coverage. A key strategy involves introducing controlled noise to the input controls, which has been shown to enhance the robustness and generalizability of the policies \cite{green1986persistence, laskey2017dart, ke2021grasping}.

\begin{figure}[!ht]
\centering
\subfloat[Without noise]{\includegraphics[height=1in]{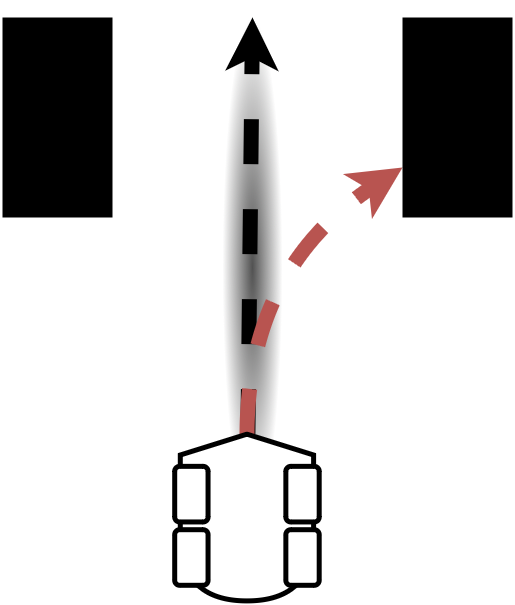}%
\label{fig:normal}}
\hfil
\subfloat[With noise]{\includegraphics[height=1in]{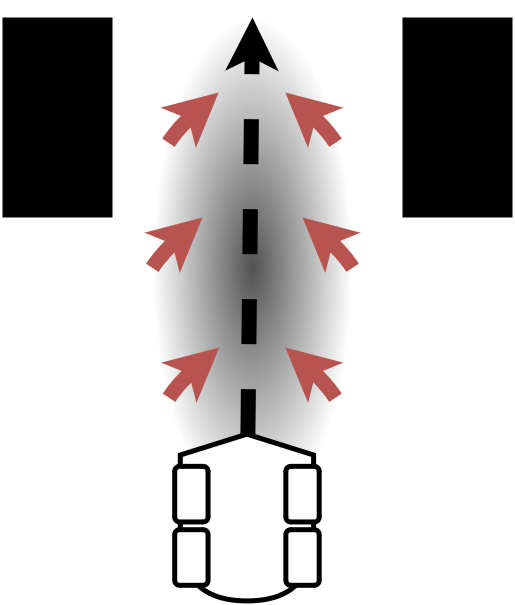}%
\label{fig:with_noise}}
\caption{Demonstrations with exploration noise allow policy to learn how to effectively act on various states}
\label{fig:noisy_demonstrations}
\end{figure}

In this letter, we introduce Learned-imitation on Cluttered Space (LiCS), an learning-based local navigation method to tackle the challenge of navigation in cluttered environments. Our approach utilizes a Transformer-based neural network for imitation learning \cite{vaswani2017attention}. During demonstrations, we inject Gaussian noise $\mathcal{N}(0, \sigma^2)$ to ensure the expert demonstrations cover a broad range of states and induce knowledge of the policy to recover from imminent collision (Fig. \ref{fig:noisy_demonstrations}). This method enables the system to learn and replicate an optimization-based baseline algorithm, adapting it to robustly and rapidly handle cluttered environments. Our proposed method demonstrates a robust capability to navigate a UGV equipped with LiDAR at speeds up to $1.5 m/s$ through narrow passages, validated in both simulation and real-world experiments.

The contribution of this paper can be listed as follows:
\begin{enumerate}
    \item Proposed an efficient approach to offline imitation learning using behavior cloning with Gaussian noise injection to input control during demonstration.
    \item Proposed a Transformer network that significantly increase the performance of imitation learning for differential drive vehicle with 2D LiDAR sensor navigating in highly cluttered space.
    \item Performed a thorough empirical study validating the robustness of the proposed learning method and neural network.
\end{enumerate}

Additionally, this work was recognized as the first-place winner in the BARN (Benchmark Autonomous Robot Navigation) Challenge at ICRA 2024 in Yokohama, Japan. Technical details about the challenge are discussed in \cite{xiao2024autonomous}.

\section{Proposed System}

\begin{figure*}[!t]
\centering
\includegraphics[width=\textwidth]{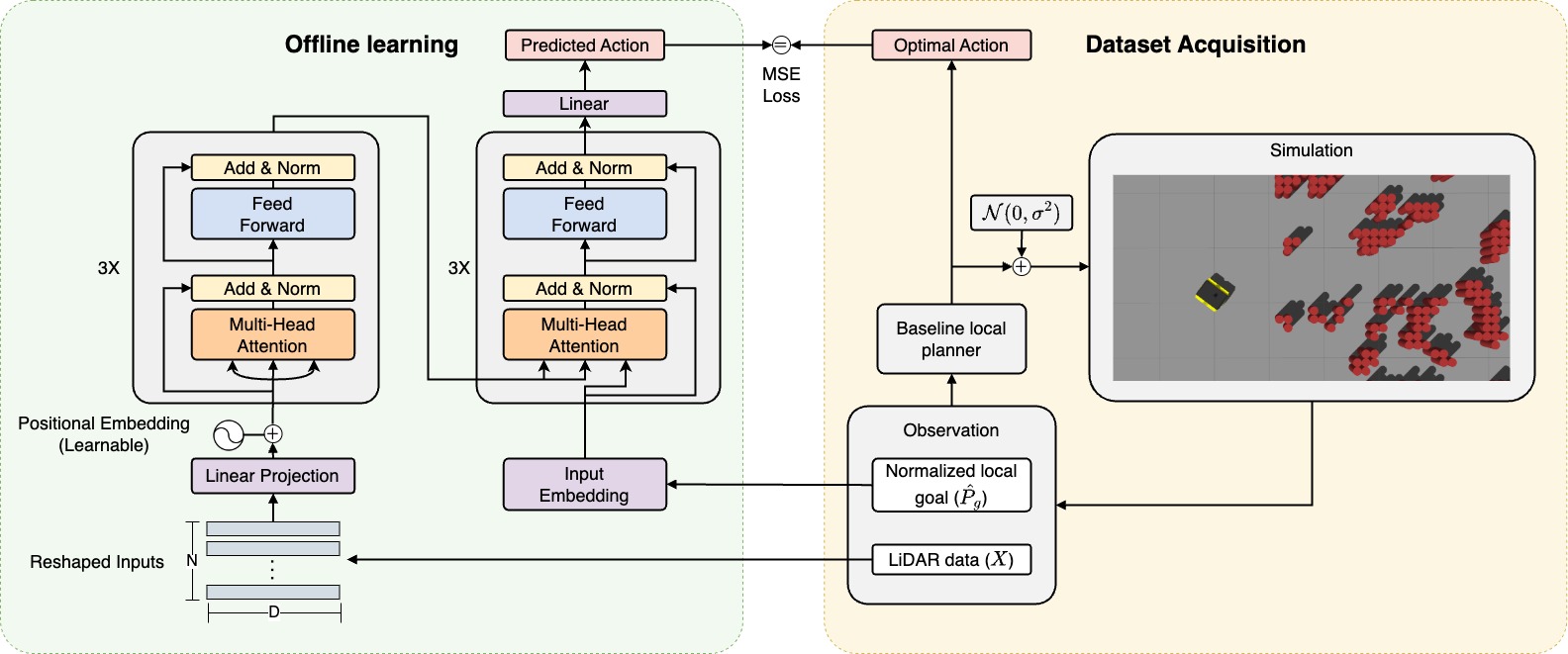}

\caption{Training pipeline diagram of the Learned-imitation on Cluttered Space model. It consists of two steps: dataset acquisition (right) and offline learning (left). During dataset acquisition, observation data and optimal action given by baseline local planner are recorded. During the training, the optimal action and predicted action by the network are compared and MSE Loss is calculated for back-propagation.}
\label{fig:training-pipeline}
\end{figure*}

This section proposes an integrated framework for training a neural network to imitate the behavior of an expert local planner for a differential drive vehicle equipped with 2D LiDAR and odometry sensors navigating in a cluttered environment. The framework consists of two essential and one optional components, an offline imitation learning pipeline and a transformer-based neural network model to learn the behavior of expert local planner effectively. In addition, a safety check layer using geometric calculations to ensure the safety of the input control during hardware implementation is proposed.

\subsection{Behavior cloning}

Given a dataset $\mathbb{D}$ consisting of sets of state-action pairs $\{(s,a^*)\}$ from simulation using an expert controller, our goal is to train a policy $\pi_\theta(s)$ with parameters $\theta$ that imitates the expert policy $\pi^*$. Behavior cloning (BC) reduces the imitation learning into a supervised learning task aimed at minimizing the following objective:

\begin{equation}
    \arg\min_\theta \mathop{\mathbb{E}_{(s,a^*)\sim \mathbb{D}}}[l(a^*,\pi_\theta(s))]
    \label{eq:bc}
\end{equation}

In our proposed system, we employ an MSE loss function to calculate the discrepancy between the expert action and the learned policy $l(a^*, \pi_\theta(s))$. The network, as shown in Fig. \ref{fig:training-pipeline}, is trained using simulation of differential robot model. We deploy the model of cluttered environment for robot navigation as proposed in \cite{perille2020benchmarking}. During training, exploration noise modeled as Gaussian function with standard deviation $\sigma$ is added to the output velocity action, modifying the input action in the simulation to include noise rather than the optimal one.

\begin{align}
    v &= v^* + \mathcal{N}(0, \sigma^2) \\
    \omega &= \omega^* + \mathcal{N}(0, \sigma^2)
\end{align}

Using the LiDAR sensor data $X$ and normalized local goal $\hat{P_g}$ as input for the neural network model, and the optimal output velocity action ($v^*, \omega^*$) provided by the baseline algorithm as the target value, we trained the proposed network in a supervised manner using MSE loss of the predicted and optimal velocity action (Eq. \ref{eq:bc}).

\subsection{Transformer-based neural network}

The network used for the imitation learning is depicted in the Fig. \ref{fig:training-pipeline}. Employing both transformer encoder and decoder, the network processes raw LiDAR sensor data and normalized local goal $\hat{P}$, derived from a global path obtained from the global planner, to provide outputs of linear and angular velocity ($v$ and $\omega$) for the robot. The LiDAR scan data is a vector with length of $H$ data, and a global path is a list of Cartesian points $P'=[(x_1, y_1), (x_2, y_2), \ldots, (x_k, y_k)]$ from the current robot position towards the global goal point.

To derive the local goal $P_g$, points in the global path are transformed from the origin $O'(0, 0)$ to the robot frame ($P_i' \rightarrow P_i$), selecting the closest point on the global path that has distance $||P_i||\geq L,\  \forall i\in 1, 2, \ldots, k$, where $L$ is the lookup distance of the robot. The normalized local goal is calculated by dividing the transformed local goal point with its magnitude.

\begin{align}
    P_g &= \operatorname{arg\,min}_{P_i : \|P_i\| \geq L} \|P_i\| \\
    \hat{P_g} &= \frac{P_g}{||P_g||}
\end{align}

The transformer encoder's architecture is derived from the ViT (Vision Transformer) \cite{dosovitskiy2021image}. As illustrated in Fig. \ref{fig:training-pipeline}, the encoder inputs raw LiDAR scan data, which consists of distance measurements obtained sequentially increasing laser angles. The input $X \in \mathbb{R}^{H \times 1} $ is first reshaped into a 2D matrix of patches $X_p \in \mathbb{R}^{N \times D}$, where $N$ is numbers of patches and $D = H/N$ is the length of the patches. Following reshaping, the input passes through a LiDAR embedding layer consisting of a linear network. After embedding, trainable positional embeddings are added. Unlike the standard ViT where class token row is added during the embedding process, we omit this step as our focus is encoding the observation data for navigation, not classification. The remainder of encoder structure mirrors that of the standard transformer \cite{vaswani2017attention}.

The transformer decoder processes the normalized local goal as input. During embedding, the input is passed through a fully connected layer to match the dimension of encoder's output. In contrast to the standard transformer, position encoding and masked multi-head attention layers are omitted due to the non-sequential nature of the input. However, the encoder-decoder attention layers are retained, allowing the network to learn the relationship between the LiDAR scan and the normalized local goal, which guides the generation of robot commands through a linear transformation.

To facilitate the deployment on embedded systems, we have minimized the number of layers in both the encoder and decoder to three, ensuring the model remains lightweight and operational on embedded devices.

\subsection{Safety check layer}

The output of the neural network is often unpredictable, especially in unexplored domains, potentially leading to unsafe behavior and collisions with obstacles. To mitigate this risk, we introduce a safety check layer that takes inputs from observation data (LiDAR or costmap generated by the global planner) and the neural network's output velocity action. The computation is performed in the sensor domain, enabling fast and efficient calculation and allowing real-time supervision of the model output before passing input control into the motor controller.

\begin{figure}[!b]
\centering
\subfloat[Linear]{\includegraphics[height=1.5in]{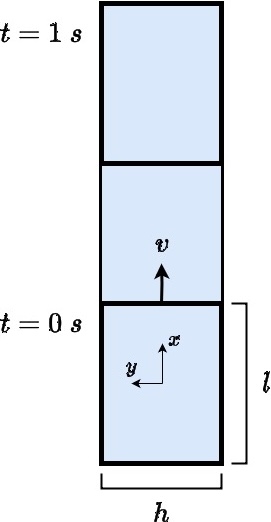}%
\label{fig:linear_case}}
\hfil
\subfloat[Radial]{\includegraphics[height=1.5in]{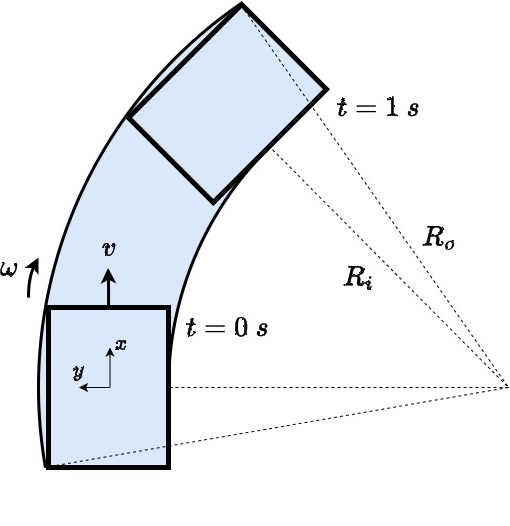}%
\label{fig:radial_case}}
\caption{ROI illustration for safety check layer during linear and radial movement.}
\label{fig:safety_check}
\end{figure}

Movement safety is predicted using geometric calculations \cite{rodrigues2022clutter}. We model the robot into a polygon centered at the middle point along the axle of the robot (illustrated in Fig. \ref{fig:safety_check}). While the shape can be arbitrary, $h$ and $l$ represents maximum distances between two points along the $y-$ and $x$-axis, respectively. For the sake of simplicity, we assume the robot's shape is a rectangle with dimensions $l$ and $h$.

\subsubsection{Linear motion with constant velocity ($|v|>0, \omega=0$)}

During linear motion, the robot moves forward or backward along the $x$-axis relative to its frame of reference, at a constant velocity ($|v|>0, \omega=0$). The safety check for linear motion involves determining whether any obstacles lie within a predefined ROI (Region of interest) directly ahead of the robot. This ROI is defined as a rectangular area extending from the front or back of the robot up to a certain distance (maximum observable distance) and covering the width of the robot.

The robot's movement is defined as "unsafe" if there exists a point with coordinate $(x, y)$ that lies withing the ROI, satisfying the following conditions:

\begin{equation}
    \begin{cases}
        xv > 0 \\
        |y| \leq h/2
    \end{cases}\label{eq:safety-linear}
\end{equation}

\subsubsection{Radial motion with constant velocity ($|v|>0, |\omega|>0$)}

During radial motion, the robot turns at a constant angular velocity ($\omega$), causing it to move along a circular arc with a turning radius $R$. The safety check for radial motion involves determining whether any obstacles lie within a predefined ROI along this arc. The turning radius $R$ is calculated as the ratio of the linear velocity ($v$) to the angular velocity ($\omega$).

Given the arbitrary width of the robot, let $R_o$ be the outer turning radius and $R_i$ be the inner turning radius. The ROI is calculated by generating two polygons of the robot at initial and final positions (after $\Delta t$) and connecting the most outer ($r=R_o$) and inner ($r=R_i$) points of the polygons with arcs. The robot's movement is defined as "unsafe" if there exists a point that lies in the ROI. For rectangle robot, we define the outer and inner radii as follows:

\begin{align}
    R_o &= \sqrt{(R+h/2)^2 + (l/2)^2} \\
    R_i &= R - h/2
\end{align}

The safety check layer can promptly detect potential collisions, enabling the system to initiate recovery actions such as reducing speed, rotating in place, or performing an emergency stop. Although this layer uses a handcrafted plan for these recovery actions, which may limit the algorithm's adaptability, it is particularly useful in real-hardware setups where safety is a priority concern.

\section{Related works}

Optimization-based methods have long been foundational in robotics for autonomous navigation, utilizing established algorithms to ensure reliable performance. The Dynamic Window Approach (DWA) \cite{fox1997dynamic} emphasizes collision avoidance by dynamically calculating optimal velocity within feasible velocity space, thus reducing computational overhead for quicker responses. Similarly, Elastic Bands (EBand) \cite{quinlan1993elastic} generate paths through a series of connected points, adjusting these paths via simulated elastic forces to avoid obstacles and minimize travel distance. Another notable optimization approach is the Free-Space Motion Tube (FSMT) technique \cite{rodrigues2022clutter}, which defines a robot's maneuvers through adaptive curvature-based motion, allowing efficient navigation in cluttered environments.

On the other hand, learning-based methods leverage advanced machine learning techniques to enhance navigational capabilities. The End-to-End (E2E) algorithm \cite{xu2023benchmarking} uses Twin Delayed Deep Deterministic Policy Gradient (TD3) to learn navigation policies directly from raw sensor inputs, allowing robots to adapt to diverse environments without predefined rules. Learning from Hallucination (LfH) \cite{xiao2021toward} simulates highly constrained obstacle configurations during training by defining a hallucination function of obstacle configuration. Then the Learning from Learned Hallucination (LfLH) \cite{wang2021agile} used self-supervised learning to generate motion plan based on the LfH and Inventec \cite{mandala2023barn} extends the LfLH by incorporating a finite state machine to manage recovery behaviors and introduces safety measures to ensure robust navigation in constrained spaces. A hybrid approach is seen in Adaptive Planner Parameter Learning from Reinforcement (APPLR) \cite{xu2021applr}, which combines traditional planning with reinforcement learning. APPLR dynamically adjusts planner parameters at each time step, merging the strengths of classical motion planning with adaptive learning to handle various navigation scenarios effectively.

These diverse approaches highlight the evolution of autonomous navigation techniques, from traditional algorithms ensuring computational efficiency to advanced learning-based methods offering adaptability and robustness, along with hybrid systems that integrate the best of both worlds.

\section{Experiment Result}

\begin{figure*}[ht!]
    \centering
    \subfloat[Success rate (higher is better)]{\includegraphics[width=0.33\textwidth]{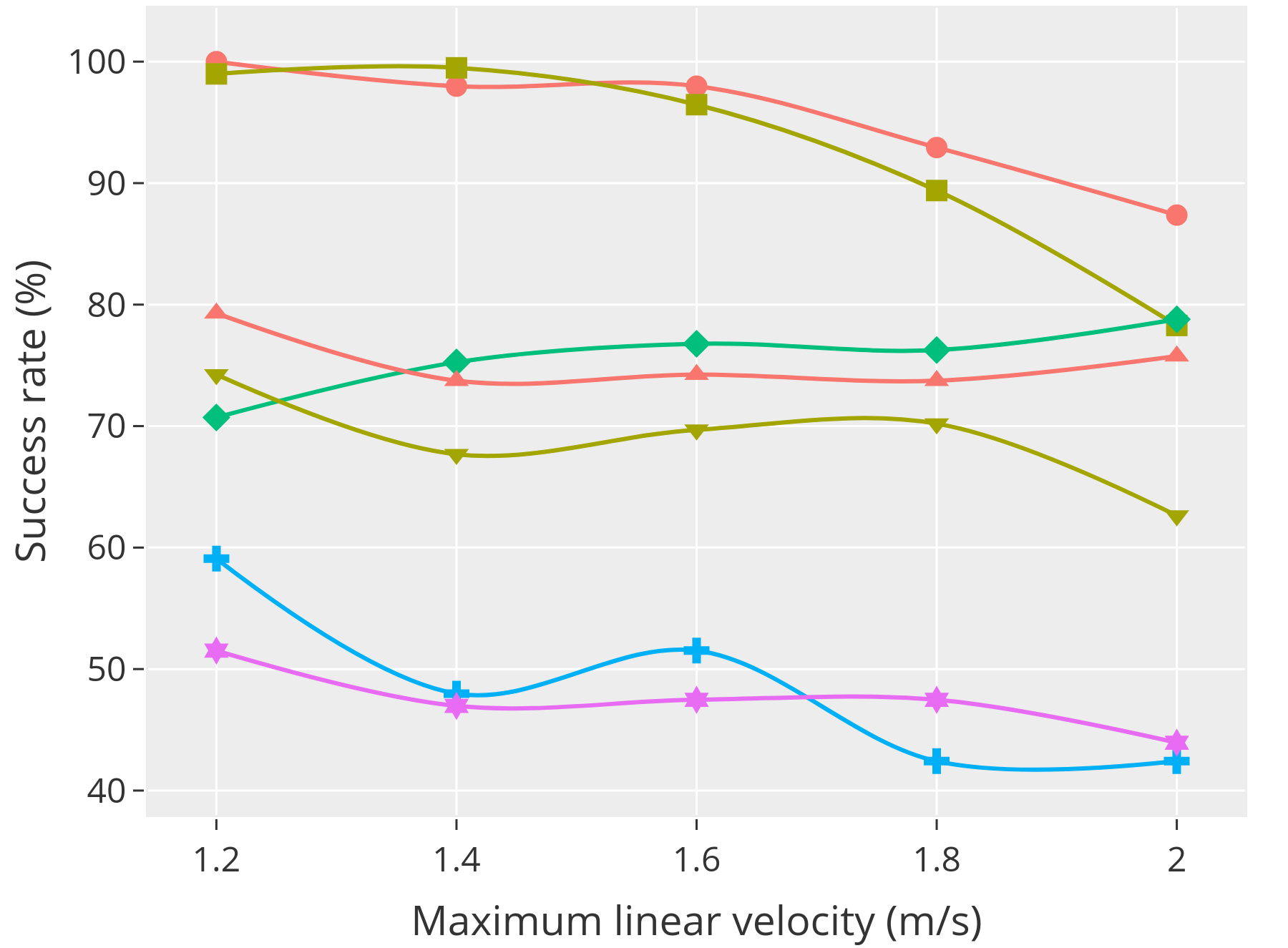}}%
    \hfill
    \subfloat[Average traversal time (lower is better)]{\includegraphics[width=0.33\textwidth]{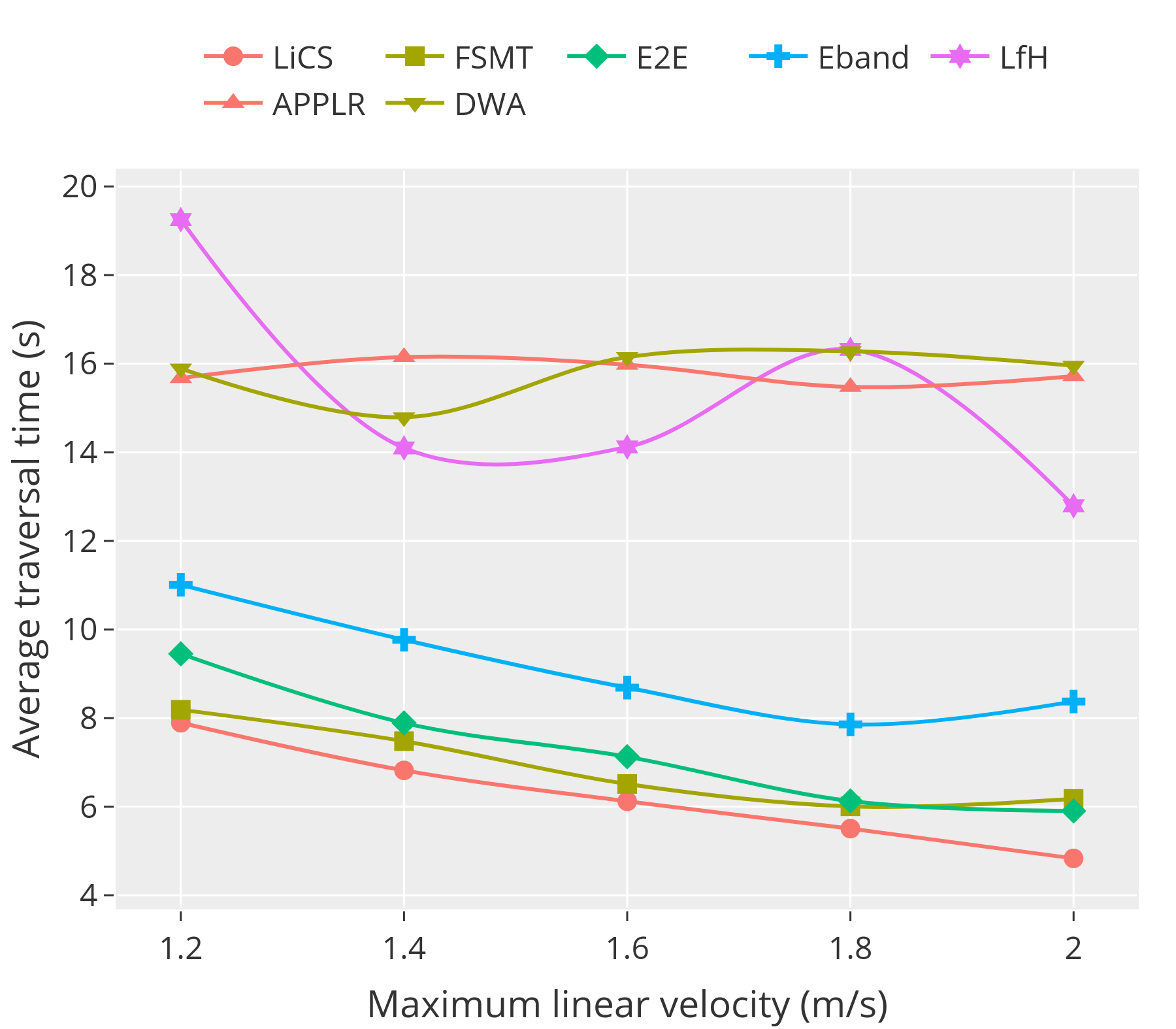}}%
    \hfill
    \subfloat[Average score (higher is better)]{\includegraphics[width=0.33\textwidth]{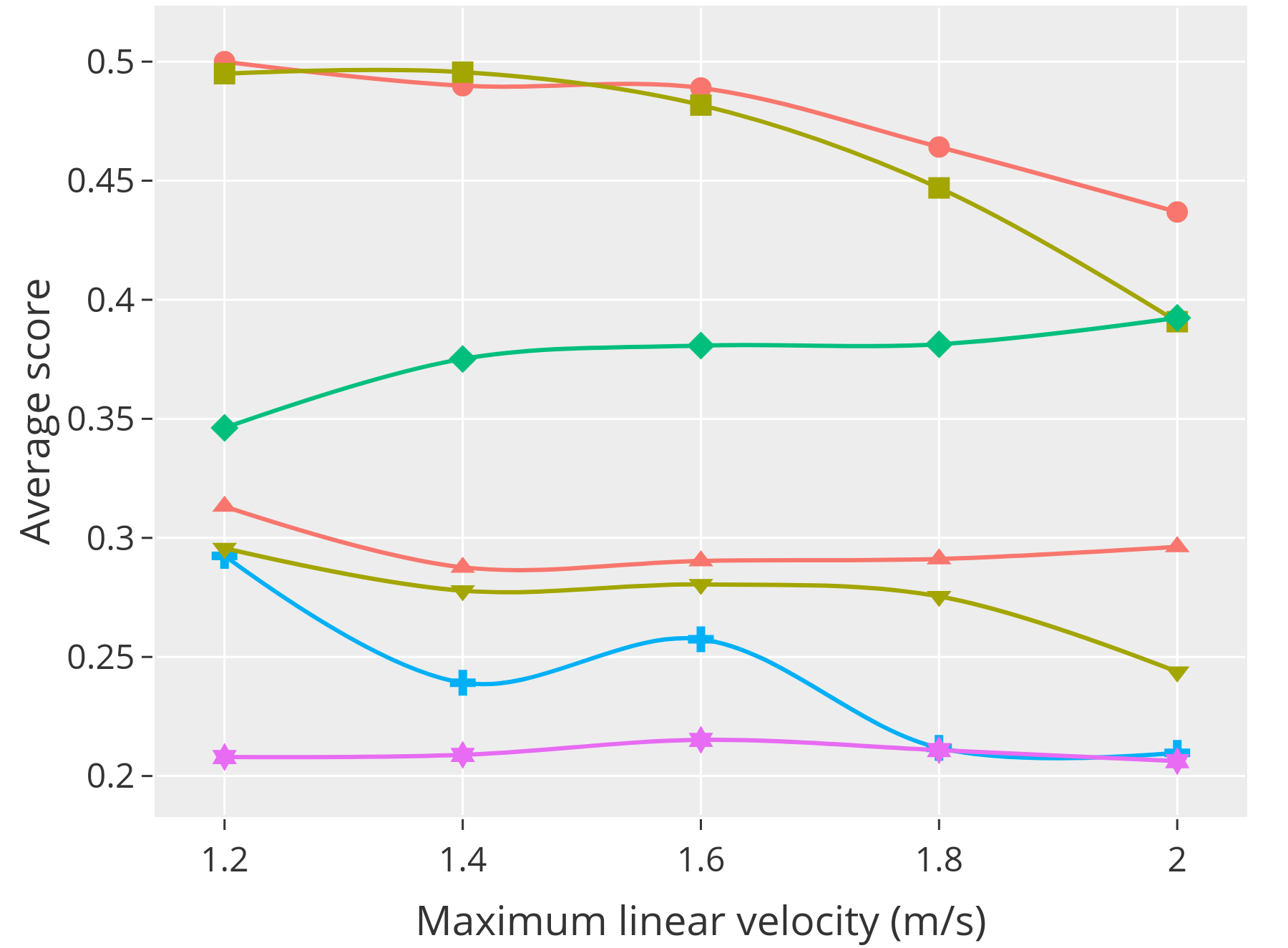}}%
    \caption{Comparison results of local navigation performance of baseline algorithms with static global planner and various maximum velocities. In this results, LiCS was implemented without safety check layer on unseen environments (benchmark worlds)}
    \label{fig:first-experiments-graph}
\end{figure*}

\begin{table*}[ht]
\centering
\caption{Performance results of first type of experiments on benchmark worlds with static global point, consisting of waypoints provided by world dataset. Each experiments on each worlds was performed in three trials.}
\label{tab:combined-metrics}
\begin{tabular}{c|ccccc|ccccc|ccccc}
\toprule
\multirow{2}{*}{\textbf{Algorithm}} & \multicolumn{5}{c|}{\textbf{Success (\%) $\uparrow$}} & \multicolumn{5}{c|}{\textbf{Avg. Time (s) $\downarrow$}} & \multicolumn{5}{c}{\textbf{Avg. Score} $\uparrow$} \\
& 1.2 & 1.4 & 1.6 & 1.8 & 2.0 & 1.2 & 1.4 & 1.6 & 1.8 & 2.0 & 1.2 & 1.4 & 1.6 & 1.8 & 2.0 \\
\midrule
DWA \cite{fox1997dynamic}     & 74.24 & 67.68 & 69.70 & 70.20 & 62.63 & 15.89 & 14.79 & 16.15 & 16.28 & 15.96 & 0.30 & 0.28 & 0.28 & 0.28 & 0.24 \\
APPLR \cite{xu2021applr}   & 79.29 & 73.74 & 74.24 & 73.74 & 75.76 & 15.68 & 16.15 & 15.98 & 15.48 & 15.72 & 0.31 & 0.29 & 0.29 & 0.29 & 0.30 \\
EBand \cite{quinlan1993elastic}   & 59.09 & 47.98 & 51.52 & 42.42 & 42.42 & 11.01 & 9.77 & 8.69 & 7.86 & 8.38 & 0.29 & 0.24 & 0.26 & 0.21 & 0.21 \\
E2E \cite{xu2023benchmarking}     & 70.71 & 75.25 & 76.77 & 76.26 & 78.79 & 9.45 & 7.89 & 7.13 & 6.13 & 5.91 & 0.35 & 0.38 & 0.38 & 0.38 & 0.39 \\
LfH \cite{wang2021agile}     & 51.52 & 46.97 & 47.47 & 47.47 & 43.94 & 19.25 & 14.10 & 14.12 & 16.32 & 12.79 & 0.21 & 0.21 & 0.22 & 0.21 & 0.21 \\
FSMT \cite{rodrigues2022clutter}    & 98.99 & \textbf{99.49} & 96.46 & 89.39 & 78.28 & 8.19 & 7.48 & 6.51 & 6.01 & 6.18 & 0.495 & \textbf{0.50} & 0.48 & 0.45 & 0.39 \\
\midrule
LiCS    & \textbf{100.00} & 97.98 & \textbf{97.98} & \textbf{92.93} & \textbf{87.37} & \textbf{7.90} & \textbf{6.82} & \textbf{6.12} & \textbf{5.51} & \textbf{4.84} & \textbf{0.500} & 0.49 & \textbf{0.49} & \textbf{0.46} & \textbf{0.44} \\
\bottomrule
\end{tabular}
\end{table*}

This section describes the implementation and evaluation of the proposed system, both in the simulation and through hardware experiments, using a differential drive UGV model with four wheels and equipped with a 2D LiDAR sensor.

\subsection{Dataset acquisition}

The data acquisition process for LiCS model involves gathering observation and optimal action data from a baseline local planner. The trial environment is Gazebo Classic simulation with ROS 1. In this step, a simulated robot navigates through various cluttered environments using two baseline control methods: FSMT (Free-Space Motion Tube) and manual control by the writer behind a PlayStation 4 joystick controller. The learned LiCS by former expert will then be referred as the base LiCS, and the later will be referred as LiCS-man.

During the recording, optimal control given by the baseline is injected using Gaussian noise with STD 0.25 before passed to the simulation. This noise allows the network to learn effectively the behavior of the baseline experts on handling various states, including forcing a near-collision states. During the trials, alongside the optimal action from baseline, observation data, including LiDAR readings and the normalized local goal, are also recorded. The parameters used for the simulation are as follow. The local goal points for each time-step is generated using A* global planner. The maximum linear and angular velocities are set to 2 $m/s$ and 3.14 $rad/s$ respectively. The line-of-sight for local goal is set to 2 $m$.

The simulation takes place across 234 different training worlds out of a total of 300, with 66 worlds reserved for benchmarking purposes. For each training environments, two successful trials without collision are saved into the database. The collected data were then used in the offline learning phase, where the model compares its predicted actions to the optimal actions using MSE loss to learn the network parameter through back-propagation.

\subsection{Simulation result}

A total 66 test worlds, referred as benchmark worlds, were used to evaluate our algorithm against the baseline methods. The proposed LiCS algorithm was trained on the remaining environments using the proposed training pipeline. The code for APPLR, EBand, E2E, LfH, and DWA was sourced from the BARN Challenge public repository \footnote{https://github.com/Daffan/the-barn-challenge}, while the code for FSMT algorithm was taken from the author's public repositories \footnote{https://github.com/romulortr/barn-kul-fm} \footnote{https://github.com/inventec-ai-center/inventec-team-barn-challenge-2023}.

Each algorithms were subjected to two types of experiments, the first is using static global planner (static local goal from starting location to the global goal, given by the dataset). This aim of this experiment is to test the pure performance of the local navigation at various maximum speed with given the identical guidance without the influence of global planner. The second experiment used A* as the global planner to test the performance of all algorithms in simulated real condition. Note that both experiments do not use localization algorithm (i.e. SLAM), hence the experiments are under influence of localization error caused by odometry drift, testing the adaptability of each local navigation algorithms.

Each trials were conducted three times. Metrics recorded include average score, success rate, and average traversal time $T$. The score metric, adapted from \cite{xia2020interactive}, incorporated traversal time as follows:

\begin{equation}
    Score = \mathds{1}_{succ} \frac{T^*}{\text{clip}(T, 2T^*, 8T^*)}
\end{equation}

The optimal traversal time $T^*$ is calculated from the shortest path length $L^*$ of each world, as provided by the dataset, divided by the maximum velocity of UGV ($2\ m/s$). Lower traversal times indicate more efficient and agile navigation. The results of these simulations are summarized in Table \ref{tab:benchmark-simulation}.

\begin{table}[ht]
\centering

\caption{Performance on benchmark worlds with A* global planner and maximum velocity 1.4 m/s. Bold and square brackets indicate the best and second best performers}
\label{tab:benchmark-simulation}

\begin{tabular}{c c c c}
\hline
\toprule
\textbf{Algorithm} & \textbf{Success (\%) $\uparrow$} & \textbf{Avg. Time (s) $\downarrow$} & \textbf{Avg. score} $\uparrow$  \\
\midrule
DWA \cite{fox1997dynamic} & 81.82  & 26.86 & 0.22  \\
APPLR \cite{xu2021applr} & 87.88 & 18.46 & 0.33 \\
EBand \cite{quinlan1993elastic} & 88.38 & 9.29 & 0.44  \\
E2E \cite{xu2023benchmarking} & 71.72 & 7.96 & 0.36 \\
LfH \cite{wang2021agile} & 97.98 & 13.11 & 0.42 \\
% Inventec \cite{mandala2023barn} & 0.4242 & [0.9899] & 13.8181 \\
FSMT \cite{rodrigues2022clutter} & [99.49] & \textbf{6.66} & [0.498] \\
\midrule
% LiCS\footnote{without safety check and dynamic inflation} & 0.4455 & 0.9091 & 8.5064 \\
% LiCS\footnote{without safety check} & [0.4874] & 0.9747& [7.8487] \\
LiCS & \textbf{100.00} & [6.85] & \textbf{0.499} \\

\bottomrule
\end{tabular}

\end{table}

\begin{table}[ht]
\centering
\caption{Performance on 16 most challenging worlds with A* global planner and maximum velocity 1.4 m/s.}
\label{table:hard result}

% \begin{tabular}{c c c c c}
% \hline
% \toprule
% \textbf{Algorithm} & \textbf{Av. score} & \textbf{Success \%} & \textbf{Collision rate} & \textbf{Av. Time}  \\
% \hline
% APPLR & 0.1151 & 0.4792 & 0.3750  & 31.65 \\
% EBand & 0.2194 & 0.6875 & 0.2917 & 19.54  \\
% E2E & 0.1250 & 0.25 & 0.7500  & 5.6100 \\
% LfH & 0.2142 & 0.5210 & 0.4792 & 14.5500 \\
% Fast-DWA & 0.1010 & 0.5210 & 0.2708 & 36.2100  \\
% FSMT & 0.2806 & 0.5625 & 0.4167 & \textbf{4.7800}  \\
% Inventec & 0.2932 & \textbf{0.9375} & \textbf{0.0625} & 21.5000 \\
% \hline
% LiCS\footnote{without safety check and dynamic inflation} & 0.3167 & 0.6458 & 0.3333 & 8.9144  \\
% LiCS\footnote{without safety check} & XXXX & XXXX& XXXX & XXXX \\
% LiCS & 0.4162 & 0.8438 & 0.1562 & 7.8345  \\

% \bottomrule
% \end{tabular}

\begin{tabular}{c c c c}
\hline
\toprule
\textbf{Algorithm} & \textbf{Success (\%)} $\uparrow$ & \textbf{Avg. Time (s)} $\downarrow$ & \textbf{Avg. score} $\uparrow$ \\
\midrule
DWA \cite{fox1997dynamic} & 58.33 & 33.45 & 0.12 \\
APPLR \cite{xu2021applr} & 52.08  & 33.45 & 0.12 \\
E2E \cite{xu2023benchmarking} & 35.42  & 9.04 & 0.18 \\
LfH \cite{xiao2021toward} & 77.08 & 18.44 & 0.27 \\

EBand \cite{quinlan1993elastic} & [70.83] & 11.48 & [0.34]  \\
% Inventec \cite{mandala2023barn} & 0.2932 & \textbf{0.9375} & 21.5000 \\
FSMT \cite{rodrigues2022clutter} & 62.50 & \textbf{7.70} & 0.31  \\
\midrule
% LiCS\footnote{without safety check and dynamic inflation} & 0.3167 & 0.6458 & 8.9144  \\
% LiCS\footnote{without safety check} & \textbf{0.4583} & [0.9167] & 8.4039 \\
LiCS & \textbf{91.67} & [7.87] & \textbf{0.46}  \\

\bottomrule
\end{tabular}

\end{table}

% \begin{figure}[!t]
%     \centering
%     \includegraphics[width=0.5\textwidth]{compare-time.png}
%     \caption{Caption}
%     \label{fig:enter-label}
% \end{figure}

% \begin{figure}[!t]
%     \centering
%     \includegraphics[width=0.5\textwidth]{compare-score.png}
%     \caption{Caption}
%     \label{fig:enter-label}
% \end{figure}

Fig. 4 demonstrates the performance comparison between our proposed LiCS model and its baseline expert, FSMT, in environments unseen during training. The experiments were conducted using a static global planner, and the results show that LiCS performs similarly to FSMT at lower maximum velocities, maintaining a comparable success rate up to 1.4 m/s. However, what is particularly noteworthy is that LiCS outperforms FSMT at higher velocities, maintaining a success rate above 80\% at 2 m/s, whereas FSMT’s performance degrades significantly at this speed.

This robust performance at higher speeds could be attributed to the noise-injection during data collection, which helped the model learn a broader range of behaviors and adapt more effectively to diverse states, including those involving rapid movements and near-collision scenarios. LiCS’s ability to handle such high speeds may also explain why it achieved a lower traversal time across all tested velocities, outperforming FSMT not just in safety but in efficiency as well.

The third best performing algorithm at higher speed, E2E, which was trained using RL (Reinforcement Learning), showed an intriguing result. Unlike other algorithms that perform best at lower speed, E2E performed best at maximum velocity of 2 $m/s$, the velocity which it is trained, inferring that RL lacks generalizability \cite{pmlr-v97-cobbe19a}.

Table \ref{tab:benchmark-simulation} and \ref{table:hard result} shows the second experiment with A* global planner. The decision to set the maximum velocity at 1.4 $m/s$ in the second experiment was influenced by the first experiment showing that all algorithms, except E2E, start to exhibit performance drops, making 1.4 $m/s$ a challenging yet manageable benchmark for comparing algorithm efficacy in realistic scenarios.

\begin{figure}[!t]
\centering
\subfloat[]{\includegraphics[width=0.45\textwidth]{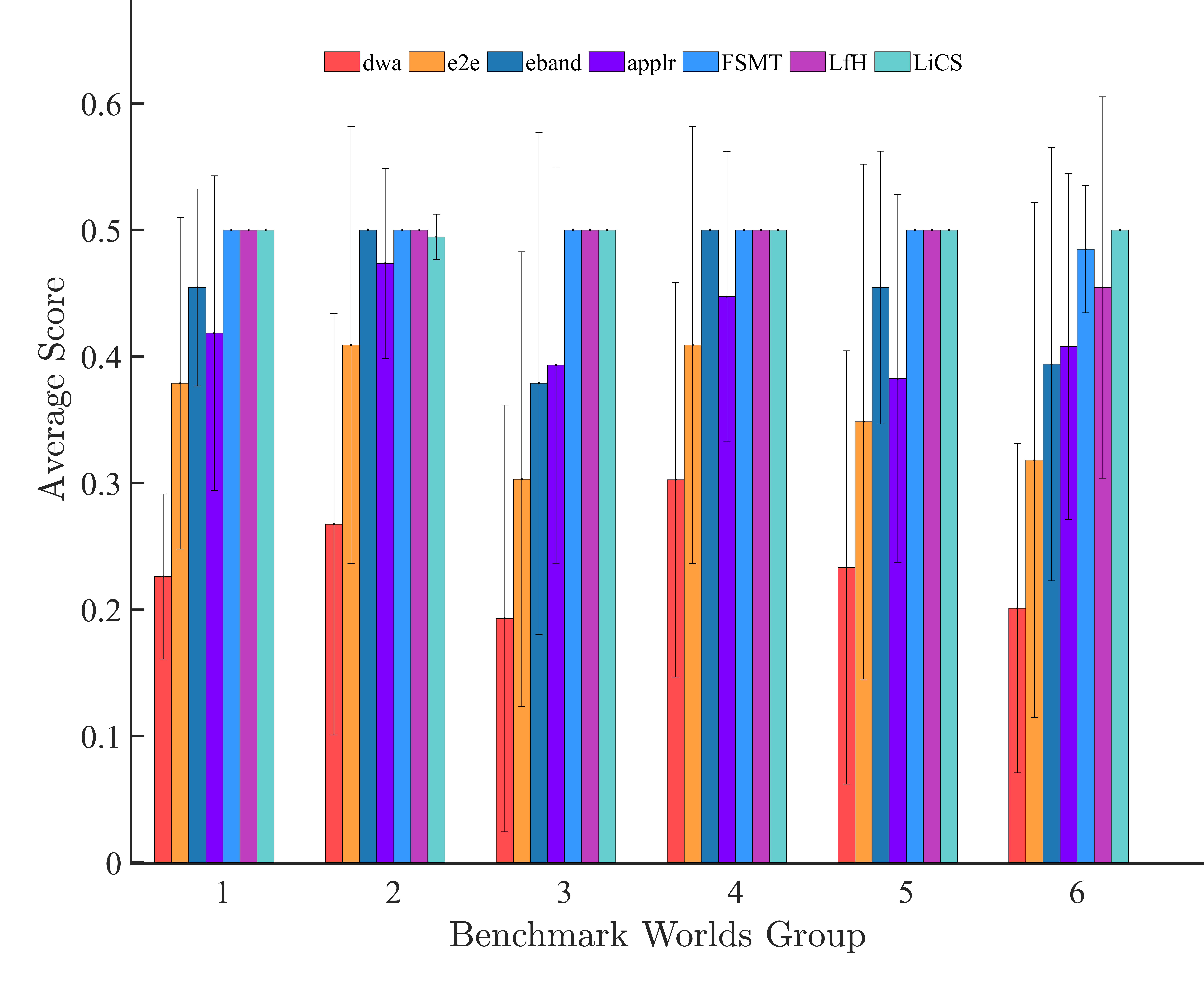}}%
\hfill
\subfloat[]{\includegraphics[width=0.45\textwidth]{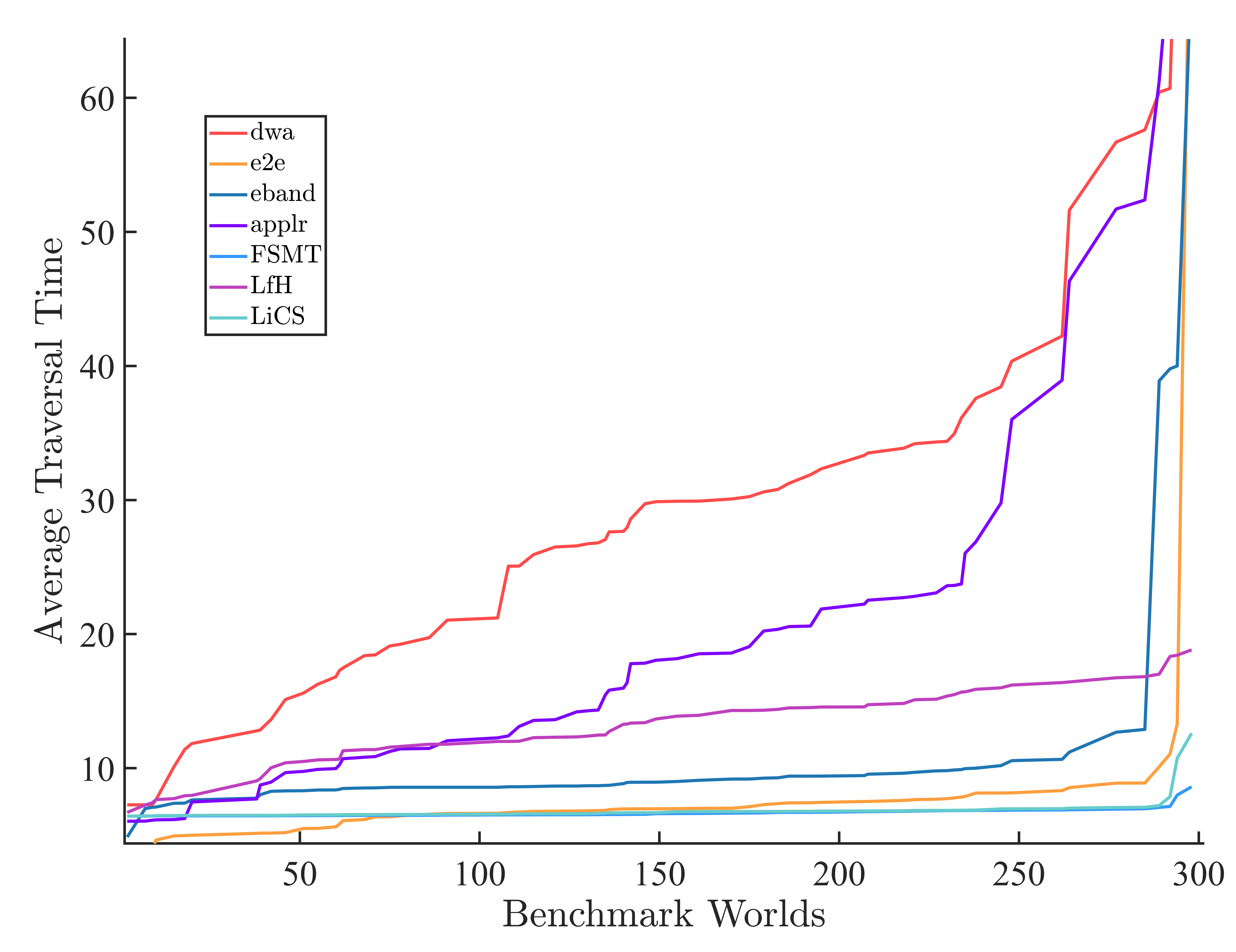}}%
\caption{(a) Average scores of algorithms for each benchmark worlds group. (b) Average traversal times across benchmark worlds.}
\label{fig:results}
\end{figure}

Our proposed algorithm, LiCS, demonstrated the highest average score, showcasing robust performance across metrics. LiCS achieved a success rate of 100.0\%. Although FSMT exhibited the shortest average traversal time of 6.6613 seconds, it suffered from a slightly lower success rate. LiCS provided a balanced approach with an average time of 6.8541 seconds, effectively combining speed and safety. To provide a clearer visualization, we plotted the average traversal times (over three trials) for each algorithm and the average score by grouping test worlds into 6 bins in Fig. \ref{fig:results} 

To evaluate the performance limits of these algorithms, we selected 16 of most challenging worlds from the BARN dataset, where most algorithms had previously scored the lowest. This selection aimed to rigorously test the robustness and adaptability of each algorithm under difficult navigation scenarios. Each algorithm was tested over three iterations in these hard worlds. The results of this evaluation are summarized in Table \ref{table:hard result}.

In these challenging environments, LiCS consistently outperformed the other algorithms, achieving the highest and second highest average scores and still maintaining high success rate. Although FSMT had the shortest average traversal time of 7.7025 seconds, it suffered a huge 36.9949\% dropout in success rate, comparing to LiCS that can maintain above 90\%. LiCS also demonstrated a rapid navigation with an average time of 7.8739 seconds.

\subsection{Ablation Study}

To evaluate the performance of the proposed Transformer model in isolation, an ablation study was conducted by replacing the LiCS model with three alternative models: MLP, RNN, and LSTM. All models were trained on the same dataset using identical parameters and then tested in benchmark environments with an A* planner and a maximum velocity of 1.4 m/s. As summarized in Table \ref{tab:ablation-result}, the success rates of all three models decreased by more than 11\% compared to the original Transformer-based model.

Furthermore, the LiCS model demonstrated solid performance when trained with a human expert. It outperformed all other baseline algorithms except FSMT. While its performance was slightly lower than both the FSMT and the LiCS model trained using FSMT as an expert, the results are still promising. Notably, the human expert involved was the first author, who possessed only basic teleoperation skills. This indicates that the model’s performance could potentially be further enhanced with training guided by a professional or robot technician.

Through the process of comparing the performance of LiCS with other algorithms, it becomes evident that our proposed algorithm strikes a balance between high success rates, low collision rates, and reasonable completion times. Unlike some algorithms that excel in one aspect but lag in others (e.g., E2E's low collision rate but poor success rate in hard worlds), LiCS maintains a strong overall performance, making it suitable for a wide range of real-world applications.
\begin{table}[ht]
\centering

\caption{Ablation study of LiCS with different NN (Neural Network) models and human expert}
\label{tab:ablation-result}

\begin{tabular}{c c c c}
\hline
\toprule
\textbf{Algorithm} & \textbf{Success (\%) $\uparrow$} & \textbf{Avg. Time (s) $\downarrow$} & \textbf{Avg. score} $\uparrow$  \\
\midrule
LiCS-RNN & 85.35 (-14.65\%) & 7.39 & 0.42 \\
% Inventec \cite{mandala2023barn} & 0.4242 & [0.9899] & 13.8181 \\
LiCS-LSTM & 85.36 (-14.64\%) & [7.32] & 0.43 \\
LiCS-MLP & 88.3 (-11.62\%) & 7.36 & 0.44 \\
LiCS-human & [98.02] (-1.98\%) & 7.78 & [0.49] \\
\midrule
% LiCS\footnote{without safety check and dynamic inflation} & 0.4455 & 0.9091 & 8.5064 \\
% LiCS\footnote{without safety check} & [0.4874] & 0.9747& [7.8487] \\
LiCS & \textbf{100.00} & \textbf{6.85} & \textbf{0.499} \\

\bottomrule
\end{tabular}

\end{table}
% \subsection{Comparison analysis}

% \begin{figure}[!t]
% \centering
% \includegraphics[width=0.5\textwidth]{success_vs_trav.png}
% \caption{Success rate vs average traversal time plot for all algorithms.}
% \label{fig:}
% \end{figure}

% Table comparing the baselines
% Percentage of completion
% Average of traversal time

\subsection{Hardware Experiment Result}

The robot used for hardware experiment is identical to the one used in simulation experiment, hence allowing the performance comparison between the simulation and real implementation (sim-to-real). The algorithm was deployed in Jackal robot equipped with Hokuyo UST-20LX LiDAR and Intel i3 CPU controller, operating without any GPU support.

We conducted tests on three tracks constructed from card boards, each offering different navigation difficulties. The first track was the easiest and the third track was the most difficult. Minimal fine-tuning was applied to the algorithm post-simulation. The identical neural model trained in the simulation was used for the hardware experiments. However the LiDAR sensor differed in resolution between the simulation ($720 \times 1$) and real hardware setups ($1081 \times 1$). To overcome this discrepancy, we scale down the hardware sensor data size by linearly sampling of the real sensor data to match the required model input size. 

\begin{figure}[!t]
\centering
\subfloat[Track I]{\includegraphics[width=0.23\textwidth]{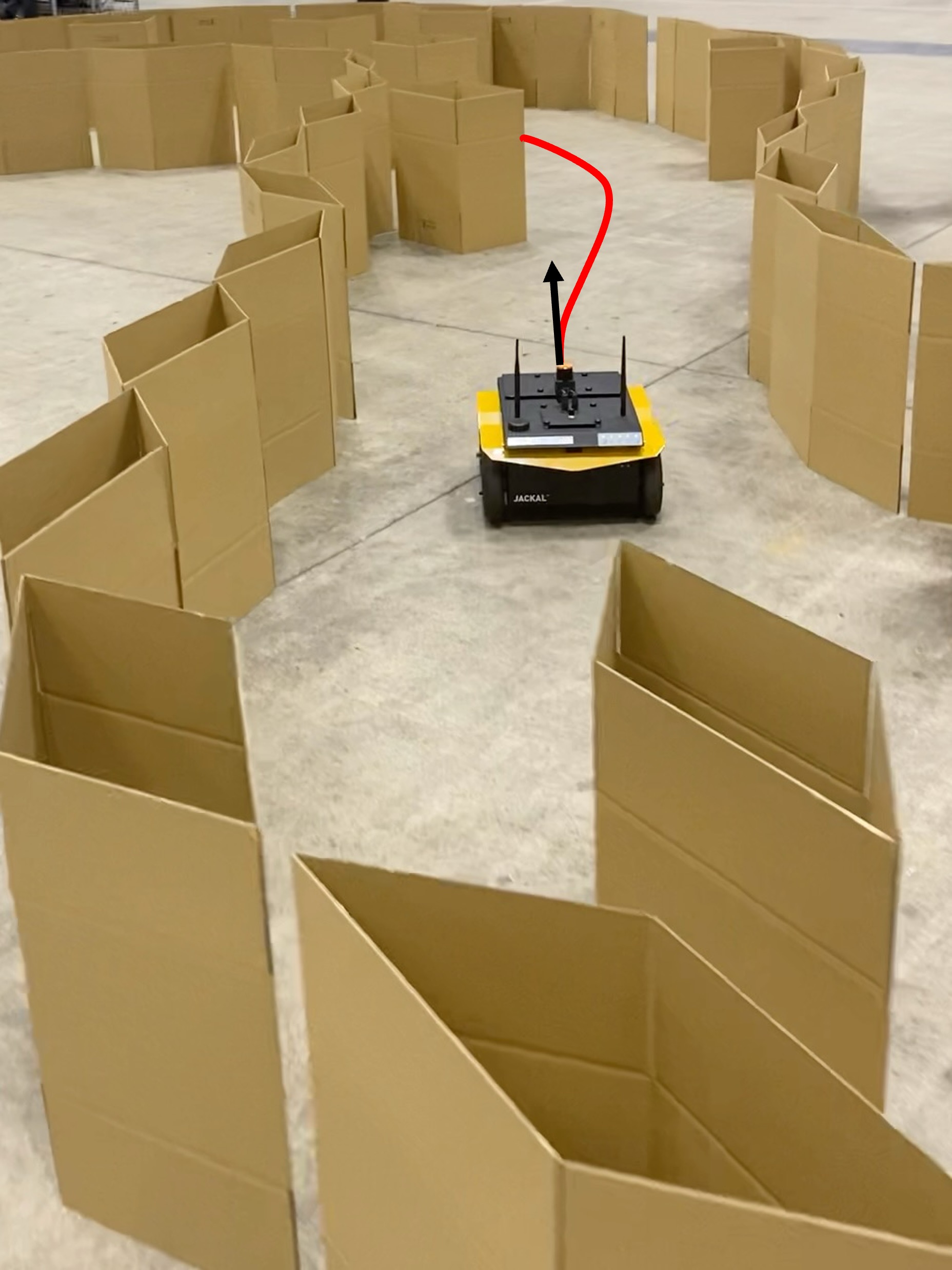}%
\label{fig:track1}}
\hfil
\subfloat[Track II]{\includegraphics[width=0.23\textwidth]{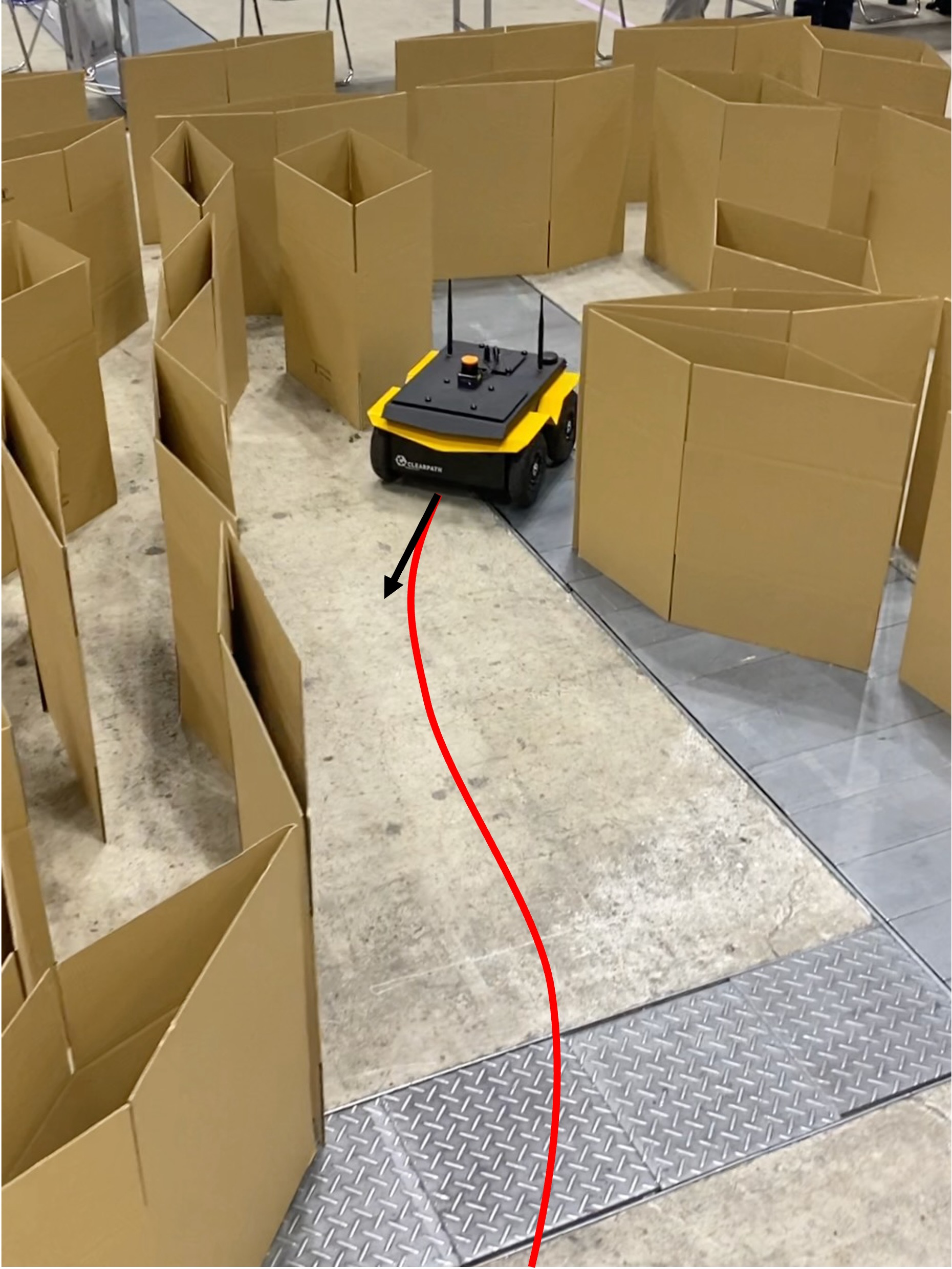}%
\label{fig:track2}}
\caption{Robots navigating the first and second tracks of BARN Challenge at ICRA 2024. The tracks are delineated using cardboards. The red lines indicate the robot's future trajectory, and the black arrows show the direction of movement.}
\label{fig:experiment_tracks}
\end{figure}

Considering that the environment is different than the one it is trained on (shown in Fig. \ref{fig:experiment_tracks}), we implement the safety check layer solely in this experiment setup.
Additional adjustments were made to the velocity settings and the inflation radius of the global cost map to optimize the performance.

\begin{table}[t!]
\centering
\caption{Traversal time required to complete trials during hardware experiments.}
\begin{tabular}{c c c c}
\hline
\toprule
\ Trial & {Track 1 (s)} & {Track 2 (s)} & {Track 3 (s)} \\
\midrule
1 & 32 & 37 & X \\
2 & 31 & 37 & X  \\
3 & 32 & 40 & X \\
4 & 27 & 29 & X \\
5 & 30 & 32 & X  \\
\midrule
Average & 30.4 & 35 & X  \\

\bottomrule
\end{tabular}
\label{table:hardward-result}
\end{table}

The results of the hardware experiments are presented in Table \ref{table:hardward-result}. The table lists traversal times for each trial, with "X" indicating a failure to complete the track caused by collision. Our algorithm successfully navigated track 1 and 2 but failed on track 3. The difficulty increased progressively across the tracks. Particularly, track 3 featured narrow path requiring sharp turns, which proved too challenging. During the experiments, several parameters were adjusted, including maximum linear velocity, angular velocity, and inflation radius. For the first three trials on tracks 1 and 2, the maximum linear and angular velocities were set to 1.0 m/s and 1.0 rad/s, respectively. In subsequent trials, the maximum linear velocity was increased to 1.5 m/s, but the optimal performance was achieved at 1.3 m/s. At higher velocities, the robot moved too aggressively, necessitating frequent corrective maneuvers, which ultimately increased traversal times. On track 3, the robot frequently collided with obstacles or became stuck at tight corners due to the demanding navigation requirements.

\section{Conclusion}

In this study, we introduced the Learned-imitation on Cluttered Space (LiCS) algorithm, a novel imitation learning-based approach for navigating Unmanned Ground Vehicles (UGVs) through complex, cluttered indoor spaces. This approach utilizes a Transformer-based neural network and combining behavior cloning with robust safety checks, LiCS was designed to optimize navigation by learning from expert demonstrations while adapting to dynamic and unpredictable conditions. It was trained under noisy conditions to generalize across various scenarios. The safety layer integrated into LiCS effectively mitigated potential hazards, preventing collisions and ensuring stable operation under diverse conditions.

The simulation result demonstrated that LiCS provides a significant improvement over baseline methods. It achieved the lowest average traversal time with high success rate, especially in challenging environments characterized by tight spaces. The hardware experiments further validated the simulation results, with LiCS performing reliably on real robots. Although it encountered difficulties in the most challenging track, which highlighted potential limitations in real-world sensor discrepancies and dynamic responses, the overall success in simpler tracks confirmed its practical utility and effectiveness. This study's findings suggest that the LiCS algorithm represents a promising advancement in the field of autonomous navigation for UGVs, particularly in scenarios where traditional methods struggle.

Furthermore, similar with other local planner algorithms, the environment used during the experiment is often assumed to be straightforward, ignoring the influence of the global planner. For a more complex tasks, such as exploration in unknown environments, a global planner (e.g., Dijkstra or A*) can be integrated into the system stack. Additionally, employing SLAM would enable simultaneous localization and mapping, which can be beneficial in environments where map-building is necessary. However, these components are not central to the current implementation and are considered optional extensions for future work.

% or


% \appendices

% % you can choose not to have a title for an appendix
% % if you want by leaving the argument blank
% \section{}

\section*{Acknowledgment}

This work was supported by Unmanned Vehicles Core Technology Research and Development Program through the National Research Foundation of Korea (NRF), Unmanned Vehicle Advanced Research Center (UVARC) funded by the Ministry of Science and ICT (MSIT), the Republic of Korea (\#2020M3C1C1A0108237512).

\ifCLASSOPTIONcaptionsoff
  \newpage
\fi



%\bibliographystyle{IEEEtran}
\bibliographystyle{IEEEtran}  % "unsrtnat" for unsorted references
\bibliography{reference} 

\begin{IEEEbiography}[{\includegraphics[width=1in,height=1.25in,clip,keepaspectratio]{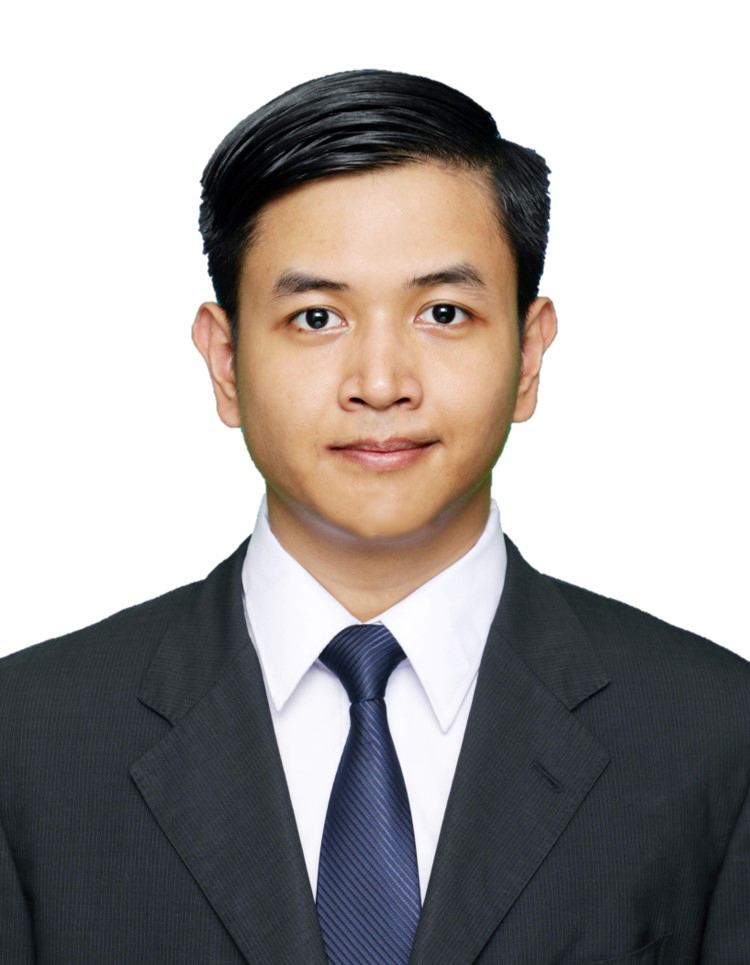}}]{Joshua Julian Damanik} received the B.S. degree in engineering physics from Institut Teknologi Bandung, Indonesia, in 2018, and the M.S. degree in aerospace engineering from the Korea Advanced Institute of Science and Technology (KAIST), Daejeon, South Korea, in 2021, where he is currently pursuing the Ph.D. degree in aerospace engineering KAIST. His current research interests include robotics learning and control, and data mining.
\end{IEEEbiography}

\begin{IEEEbiography}[{\includegraphics[width=1in,height=1.25in,clip,keepaspectratio]{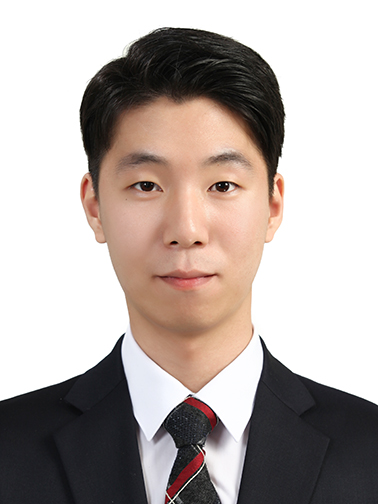}}]{Jae-Won Jung} received the B.S. degree in mechanical engineering from SungKyunKwan University, Suwon, South Korea in 2018. He is currently pursuing the M.S. degree in aerospace engineering from the Korea Advanced Institute of Science and Technology (KAIST), Daejeon, South Korea. His current research interests include robotics control, and Machine Learning.
\end{IEEEbiography}
\begin{IEEEbiography}[{\includegraphics[width=1in,height=1.25in,clip,keepaspectratio]{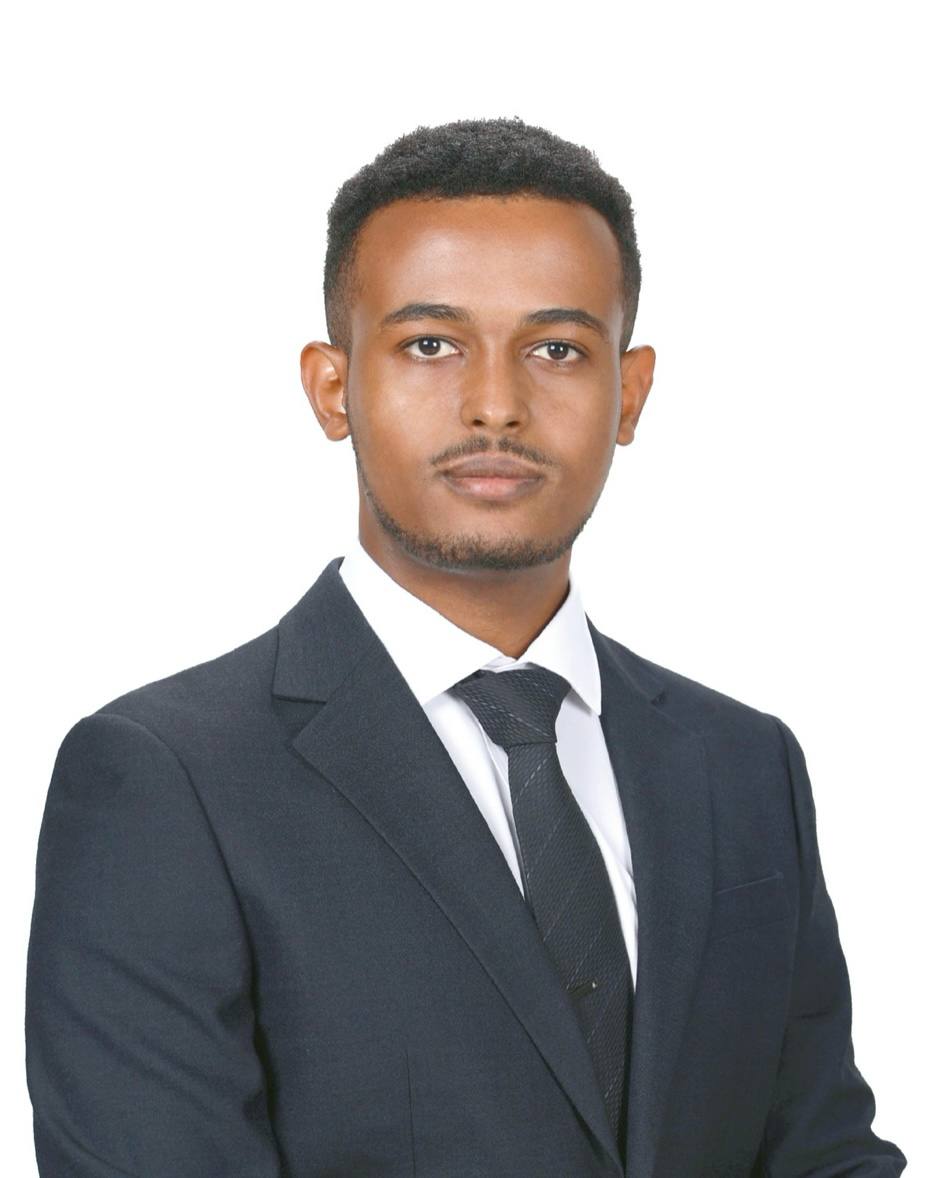}}]{Chala Adane Deresa} received the B.S. degree in aerospace engineering from the Korea Advanced Institute of Science and Technology (KAIST), Daejeon, South Korea, in 2024, where he is currently pursuing the M.S. degree in aerospace engineering. His current research interests include robotics estimation and control, and spacecraft autonomy.
\end{IEEEbiography}
%\newpage
\begin{IEEEbiography}[{\includegraphics[width=1in,height=1.25in,clip,keepaspectratio]{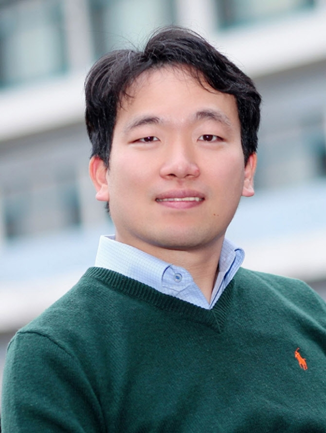}}]{Han-Lim Choi} (Senior Member, IEEE) received the B.S. and M.S. degrees in aerospace engineering from the Korea Advanced Institute of Science and Technology (KAIST), Daejeon, South Korea, in 2000 and 2002, respectively, and the Ph.D. degree in aeronautics and astronautics from the Massachusetts Institute of Technology (MIT), Cambridge, MA, USA, in 2009. Then, he studied at MIT as a Postdoctoral Associate until he joined KAIST, in 2010. He is currently a Professor of aerospace engineering at KAIST. His current research interests include estimation and control for sensor networks and decision making for multi-agent systems. He was a recipient of the Automatic Applications Prize, in 2011 (together with Dr. Jonathan P. How).
\end{IEEEbiography}


\vfill



\end{document}